# OPENNDD: OPEN SET RECOGNITION FOR NEURODEVELOPMENTAL DISORDERS DETECTION

*Jiaming Yu[1], Zihao Guan[1], Xinyue Chang[1], Shujie Liu[1], Zhenshan Shi[2], Xiumei Liu[3], Changcai Yang[1], Riqing Chen[1], Lanyan Xue[1], Lifang Wei[1,*]*

[1]College of Computer and Information Science, Fujian Agriculture and Forestry University, China.
[2]Department of Radiology, the First Affiliated Hospital of Fujian Medical University, China.
[3]Developmental and Behavior pediatrics Department, Fujian Children's Hospital, China.

## ABSTRACT

Since the strong comorbid similarity in NDDs, such as attention-deficit hyperactivity disorder, can interfere with the accurate diagnosis of autism spectrum disorder (ASD), identifying unknown classes is extremely crucial and challenging from NDDs. We design a novel open set recognition framework for ASD-aided diagnosis (OpenNDD), which trains a model by combining autoencoder and adversarial reciprocal points learning to distinguish in-distribution and out-of-distribution categories as well as identify ASD accurately. Considering the strong similarities between NDDs, we present a joint scaling method by Min-Max scaling combined with Standardization (MMS) to increase the differences between classes for better distinguishing unknown NDDs. We conduct the experiments in the hybrid datasets from Autism Brain Imaging Data Exchange I (ABIDE I) and THE ADHD-200 SAMPLE (ADHD-200) with 791 samples from four sites and the results demonstrate the superiority on various metrics. Our OpenNDD achieves promising performance, where the accuracy is 77.38%, AUROC is 75.53% and the open set classification rate is as high as 59.43%.

## 1. INTRODUCTION

Autism spectrum disorder (ASD) is a disorder of the nervous system that results in difficulties in speech, social interaction and communication deficits, repetitive behaviors, and delays in motor abilities [1]. It is one of the types of neurodevelopmental disorders (NDDs). NDDs are a group of early-onset disorders affecting brain development and function [2]. It includes ASD, disorder of intellectual development (DID), attention-deficit hyperactivity disorder (ADHD), developmental speech or language disorder (DSD), specific learning disorders, and motor disorders among others [3]. Evidences have shown that NDDs have overlapping phenotypes, frequently co-occur, and share multiple genetic causes, indicating strong similarities between them [4]–[6]. In general, the ASD-aided diagnosis model in existing research can solely distinguish between ASD patients and typical developing (TD) individuals [7]–[9]. However, regarding NDDs beyond ASD, which means those categories that have not been encountered in the training phase, the ASD-aided diagnosis model is incapable of ruling them out. Due to the strong similarities between NDDs, it is highly likely that these NDDs will be diagnosed by the model as patients with ASD when they are not. This can result in these patients missing crucial early intervention, severely impacting the control, management, and care of their condition.

To address the above issue, we introduce open set recognition (OSR) for the ASD-aided diagnosis. The closed set setting assumes that all testing classes are known in the training. However, in the real world, the knowledge of classes is limited, and unknown classes may be introduced into an algorithm during the testing phase [10]. Consequently, OSR goes beyond the closed set setting and constructs a robust recognition system that identifies test samples as known or unknown, and correctly classifies all known test samples. The traditional ASD-aided diagnosis model operates under a closed set setting. For categories that are not included in the training phase, such as ADHD, DSD, or DID, we need to classify them as a single category called "unknown".

Inspired by the above observations, we design an open set recognition method based on Adversarial Reciprocal Points Learning (ARPL) [10] for ASD-aided diagnosis to alleviate misdiagnosis among similar NDDs. It is the first application of OSR in the field of ASD-aided diagnosis. To broaden the distinction between known classes in in-distribution (ID) and unknown classes in out-of-distribution (OOD), we propose a joint scaling method by Min-Max scaling combined with Standardization (MMS), which significantly improves the differences between the ID and OOD data. And the

* Corresponding author

Maximum Mean Discrepancy (MMD), a domain adaptation method, is used to make a distinction between TD subjects and ASD subjects by reducing the variability among TD subjects. Moreover, we design a Reciprocal Opposition Experiment (ROE) to verify the feasibility and robustness of our proposed method. The same four sites of open hybrid datasets from Autism Brain Imaging Data Exchange I (ABIDE I) and THE ADHD-200 SAMPLE (ADHD-200) are used for experimental verification and evaluation to demonstrate the superiority on various metrics.

## 2. METHODOLOGY

In this section, we first introduce an overview of our proposed framework. Then, we detail each part of the proposed OpenNDD.

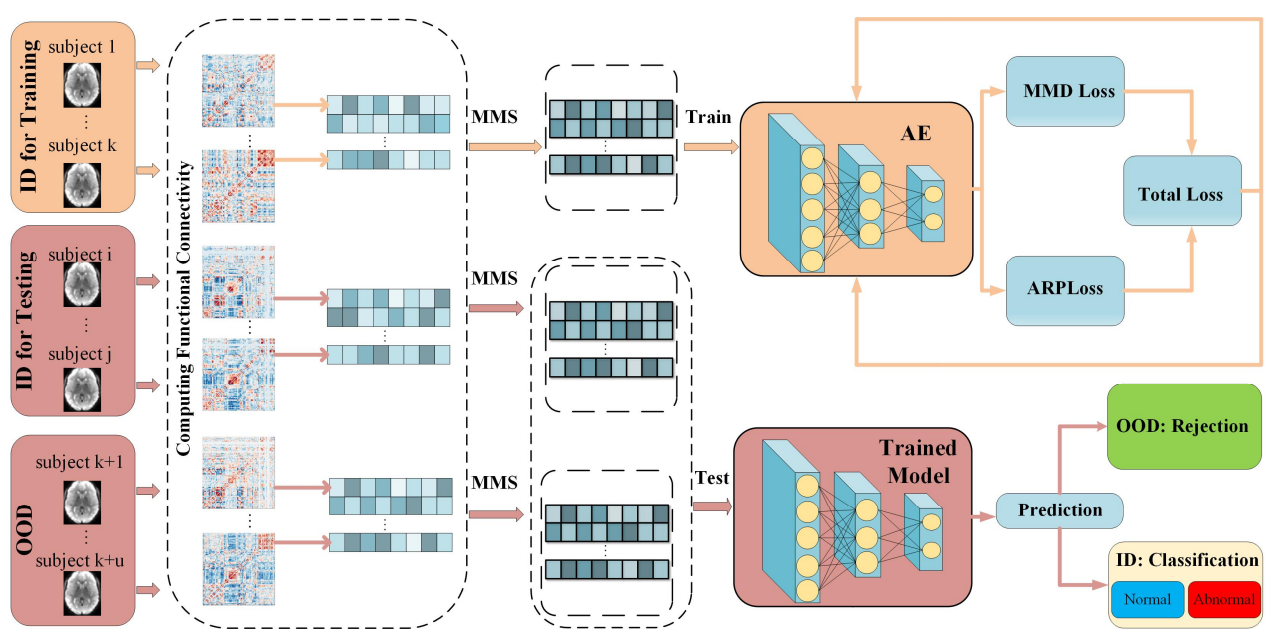

**Fig. 1.** The framework of OpenNDD.

### 2.1. Overview

The overview of the architecture is given in Fig. 1. We extract the mean time series for each region of interest (ROI) corresponding to the Anatomical Automatic Labeling (AAL) atlas from rs-fMRI, which is subsequently utilized in the construction of functional connectivity (FC). Then, we employ MMS to enlarge the differences between ID and OOD data. In the training stage, ID subjects are subsequently trained with autoencoder to obtain feature embedding. In addition, a domain adaptation strategy MMD is applied to reduce the differences between TD subjects from ABIDE I and ADHD-200. And the total loss is composed of MMD loss and ARPL loss. In the testing stage, the input data are different from the training stage, which includes the ID subjects and OOD subjects.

### 2.2. Open set recognition for NDDs detection

Given a training set of $n$ subjects $D_L = \{(x_1, y_1), \dots, (x_n, y_n)\}$ includes $K$ known categories, where $y_i \in \{1, \dots, K\}$ represents the category of $x_i$. There are $m$ test subjects in testing set $D_T = \{t_1, \dots, t_m\}$ where $t_i$ belongs to categories $\{1, \dots, K, K+1, \dots, K+U\}$ , and $U$ is the number of unknown categories.

We have two categories in ID includes ASD and TD. Inspired by [10], for a certain category $k$ in ID, we propose the concept of reciprocal point $P^k$ as the potential representation of category $k$. We denote the representation space of category $k$ as $S_k$ and its corresponding open space is designated as $O_k$. Hence, the subjects in $O_k$ should be closer to the $P^k$ than the subjects in $S_k$. Based on the nature of reciprocal point, we can classify the subjects as ASD and TD. If a sample is further from the reciprocal point of category $k$, it is more likely to be assigned to category $k$. Finally, the softmax function is applied to normalize the classification probability.

$$L_c(x; \theta, P) = -\log p(y = k | x, C, P) \quad (1)$$

Where $p$ denotes the final classification probability. In addition to reduce the empirical classification risk of the TD and ASD in ID, we introduce the adversarial margin constraint [10] to reduce open space risk from NDDs beyond ASD. For the separation of $S_k$ and $O_k$ to a larger extent, the open space $O_k$ has to be limited so that the space of open set can be confirmed. Our goal is to reduce the open space risk of each known category by limiting the open space $O_k$ to a finite range. Obviously, it is almost impossible to govern open space risk by limiting open space as there are a great number of unknown subjects in $D_U$. Taking into account spaces $S_k$ and $O_k$ are mutually complementary, the open space risk can be constrained indirectly by restricting the distance between the subjects from $S_k$ and the $P^k$ to be less than $R$ as follows:

$$L_o(x; \theta, P^k, R^k) = \max(d_e(C(x), P^k) - R, 0) \quad (2)$$

Where $R$ is a learnable margin to obtain more non-*k* subjects and $d_e$ denotes the Euclidean distance function. In such multicategory interactions, the known categories are restrained each other. Each known category is maximally forced to the margin of limited feature space to keep each category away from its potential unknown space.

Furthermore, due to the distribution shift from single class in the ADHD-200 causes a significant drop in classification accuracy, we incorporate Maximum Mean Discrepancy (MMD) [11] into our framework. Specifically, TD and ASD are labeled as ‘0’ and ‘1’ in our experiments, respectively. However, there are two subtypes that caused the distribution shift in the samples of TD. We refer to subtype 1 as the source domain and subtype 2 as the target domain. The MMD is to find a mapping function that minimizes the distance between the transformed source domain data and the target domain data.

Consequently, the final loss is constructed as follows:

$$L_{final} = L_c + \lambda_o L_o + \lambda_{mmd} L_{mmd} \tag{3}$$

Where $\lambda_o$ and $\lambda_{mmd}$ denote the coefficient of $L_o$ and $L_{mmd}$, respectively.

### 2.3. Joint Min-Max Scaling and Standardization (MMS)

Since NDDs exhibit strong clinical similarities, it is challenging to differentiate data between ID and OOD. While functional connectivity (FC) is integrated into AE for training and prediction in ASD diagnosis, conventional classifiers struggle to differentiate between ID and OOD data, primarily due to the influence of OOD data. To address this limitation, we perform MMS on FC as follows:

$$\begin{gathered} M_{min-max} = a + \frac{(M - M_{min})(b-a)}{M_{max} - M_{min}} \\ D = \frac{M_{min-max} - \bar{M}_{min-max}}{M_{std}} \end{gathered} \tag{4}$$

where $M$ is a matrix of FC and its value is to be mapped into the interval $[a, b]$. Applying Min-Max scaling to $M$, we can get a new distribution of all matrixes which is $\bar{M}_{min-max}$ and we set $a = -1$ and $b = 1$. Subsequently, the new feature space $D$ is calculated, where $D$ follows the standard Normal distribution.

**Table 1.** Quantitative evaluation of different open set recognition methods with same setting.

| Model | ACC | AUROC | OSCR | SPE | SEN | AUIN | AUOUT |
|---|---|---|---|---|---|---|---|
| Softmax-based | 78.16±5.78 | 73.23±4.47 | 58.37±5.77 | 68.22±7.60 | 88.96±6.50 | 83.00±3.33 | 54.68±6.11 |
| GCPL-based | 76.05±4.68 | 73.29±6.76 | 58.85±6.14 | 67.65±8.97 | 85.35±7.14 | 82.86±4.73 | 55.24±9.12 |
| RPL-based | 78.05±4.76 | 73.51±7.14 | 59.31±7.20 | 68.48±6.59 | 88.42±5.66 | 81.76±5.38 | 56.40±9.37 |
| OpenNDD | 77.38±5.92 | 75.53±6.01 | 59.43±6.98 | 66.35±8.69 | 89.51±6.41 | 83.83±4.71 | 57.77±7.54 |

## 3. EXPERIMENTAL SETTINGS

### 3.1. Dataset and Experimental Details

We evaluate our framework on the multi-site hybrid fMRI datasets from ABIDE I (http://fcon_1000.projects.nitrc.org/indi/abide/) and ADHD-200 (http://fcon_1000.projects.nitrc.org/indi/adhd200/), which are both preprocessed by NeuroImaging Analysis Kit (NIAK) pipeline in Preprocessed Connectomes Project (PCP) [12]. We select rs-fMRI data after preprocessing from four sites (KKI, NYU, PITT, OHSU) that are common to both datasets as our experimental data for the purpose of reducing the impact of domain shifts, which consists of 791 subjects including 144 ASD, 177 ADHD and 470 TD. Our experimental setup as follows: 470 TD subjects are divided into three groups to maintain a balanced 1:1:1 ratio of TD, ASD, and ADHD subjects. We conduct a 5-fold cross-validation experiment involving ASD and ADHD for each TD group, resulting in a total of 15 experimental results. The final experimental results represent the averages of these 15 individual results.

We evaluate the performance of our framework on seven metrics: accuracy (ACC), AUROC, open set classification rate (OSCR), specificity (SPE), sensitivity (SEN), AUIN and AUOUT [10], [13], where AUROC (ability to distinguish ID and OOD data), OSCR (ability of open set classification), AUIN (ability to distinguish ID data) and AUOUT (ability to distinguish OOD data) are metrics of OSR. With regard to implementation details, the architecture of autoencoder is configured as follows: It has an input dimensionality of 6670 features and produces an output consisting of two softmax probability values. The architecture comprises three hidden layers with 2000, 1000, and 500 units, respectively. We run the model with 16 batch size over 100 epochs. The momentum stochastic gradient descent (Momentum SGD) optimizer is used for classifier training [14]. The learning rate of the classifier starts at 0.01 and decreases by a factor of 0.1 with every 30 epochs over the course of training progress.

### 3.2. Experimental Results

The quantitative evaluation of different open set recognition methods with same setting are shown in **Table 1**. The optimal results are indicated in red. The models in **Table 1**, except for OpenNDD, indicate that we substituted ARPL in OpenNDD with other open set recognition methods. Our proposed OpenNDD using only known training samples significantly outperforms other models with other OSR methods. Specifically, OpenNDD surpasses other models by more than 2% in terms of AUROC. It indicates that OpenNDD effectively distinguish between in-distribution and out-of-distribution samples, and has superior performance in detecting previously unidentified NDDs diseases.

**Table 2.** Quantitative evaluation of the proposed crucial modules in OpenNDD.

| MMS | MMD | ACC | AUROC | OSCR | SPE | SEN | AUIN | AUOUT |
|---|---|---|---|---|---|---|---|---|
| | | 76.27±3.94 | 17.88±4.63 | 18.11±4.69 | 69.90±8.54 | 83.26±7.17 | 46.13±1.83 | 23.79±1.08 |
| ✓ | | 76.84±4.92 | 63.31±6.87 | 52.41±6.92 | 69.12±8.72 | 85.13±4.14 | 75.25±5.12 | 46.21±6.26 |
| ✓ | ✓ | 77.38±5.92 | 75.53±6.01 | 59.43±6.98 | 66.35±8.69 | 89.51±6.41 | 83.83±4.71 | 57.77±7.54 |

We further conduct ablation studies to evaluate the effectiveness of the crucial modules in OpenNDD. As documented in **Table 2**. Our observations indicate that OpenNDD struggles to differentiate out-of-distribution samples without MMS and MMD, despite performing better in in-distribution classification. OpenNDD is limited in categorizing sample types only as ASD or TD, without the ability to exclude categories beyond these two. The MMS module is introduced to address this limitation and enhance the differentiation between in-distribution and out-of-distribution samples. The results indicate that our OpenNDD with MMS can achieve 45.5%, 34.3%, 29.1%, and 22.5% improvements in AUROC, OSCR, AUIN, and AUOUT compared to OpenNDD without MMS, respectively. By introducing the MMD, OpenNDD demonstrates noteworthy enhancements across all metrics, with a slight decrease in SPE. The observations reveal that the MMD module enhances the boundary of the in-distribution and out-of-distribution by performing domain adaptation approach for TD subjects, resulting in a rise in metrics like AUROC.

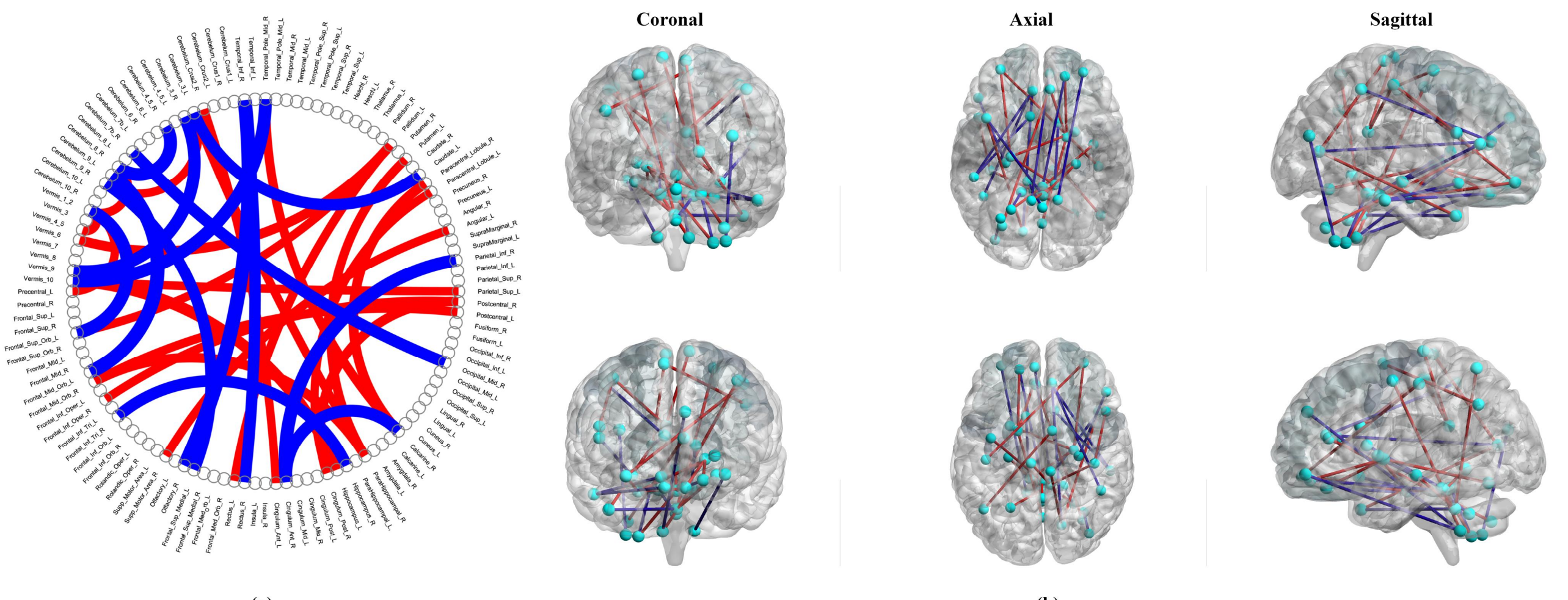

**Fig. 2.** Model interpretability. (a) Visualization of all 34 selected functional connections in the circus plot. (b) All 34 functional connections presented on the brain cortical surface. The red and blue lines in (a) and (b) depict stronger and weaker functional connections in ASD compared to TD, respectively.

**Table 3.** Effectiveness of the proposed crucial modules in ROE.

| MMS | MMD | ACC | AUROC | OSCR | SPE | SEN | AUIN | AUOUT |
|---|---|---|---|---|---|---|---|---|
| | | 72.93±5.23 | 20.48±8.61 | 18.26±7.40 | 66.65±10.74 | 78.31±6.62 | 55.77±4.73 | 19.63±1.72 |
| ✓ | | 72.63±4.80 | 71.70±6.62 | 56.02±6.49 | 64.00±7.02 | 80.35±8.64 | 86.41±3.19 | 45.52±8.85 |
| ✓ | ✓ | 73.23±5.22 | 74.95±5.69 | 57.02±6.31 | 63.65±7.34 | 81.87±7.25 | 86.54±3.78 | 51.55±7.58 |

To verify the feasibility and robustness of our proposed method, we conduct the ROE experiments, utilizing ADHD and TD as in-distribution data and ASD as out-of-distribution data. The experimental procedure aligns with the methodology outlined in **Table 2**. The consistency trend of the results shown in **Table 3** and **Table 2** further demonstrates and supports the robustness of our proposed method.

### 3.3. Model Interpretability

For model interpretation, we employ the gradient-based saliency map method [15] to identify important brain regions and connections that play a significant role in the classification process. We visualize the 34 (top 0.5%) connections with the highest absolute values among all 6670 connections in the saliency map, along with their corresponding brain regions. This visualization is presented in Fig. 2(a). Fig. 2(b) depicts the distribution of the selected 34 functional connections on the brain cortical surface by BrainNet Viewer [16]. Specifically, the brain regions identified by saliency map include Middle Temporal Gyrus (MTG), Paracingulate Gyrus, Cingulate Gyrus, demonstrate strong discriminative ability. These findings are in line with the previous studies on neuroimaging biomarkers and pathological investigations of ASD [17]–[19].

## 4. CONCLUSION

In this paper, we propose OpenNDD framework, which is the first application of OSR for ASD-aided diagnosis. The experimental results prove that OpenNDD can distinguish known and unknown categories as well as identify ASD accurately. ROE results also demonstrate that OpenNDD exhibits superior robustness and feasibility. Moreover, we show that OpenNDD identifies brain functional connectivity that contributes to ASD classification. The proposed method is of great significance for the clinical diagnosis of ASD and even for all NDDs.